\documentclass[conference]{IEEEtran}
\IEEEoverridecommandlockouts
\usepackage{graphics} %
\usepackage[pdf]{graphviz}
\usepackage{multirow}
\usepackage[font=footnotesize]{caption}
\usepackage[font=footnotesize]{subcaption}

\usepackage{xpatch}
\makeatletter
\newcommand*{\addFileDependency}[1]{%
  \typeout{(#1)}
  \@addtofilelist{#1}
  \IfFileExists{#1}{}{\typeout{No file #1.}}
}
\makeatother
\xpretocmd{\digraph}{\addFileDependency{#2.dot}}{}{}

\usepackage{mathtools}
\usepackage[%
    group-minimum-digits=4,     %
    per-mode=symbol,            %
]{siunitx}

\usepackage[T1]{fontenc}

\begin{document}

\title{\LARGE \bf
Modular Fault Diagnosis Framework\\ for Complex Autonomous Driving Systems
}

\author{Stefan Orf$^1$, Sven Ochs$^1$, Jens Doll$^1$, Albert Schotschneider$^1$, Marc Heinrich$^1$, Marc Ren\'{e} Zofka$^{1}$, \\J. Marius Zöllner$^{1,2}$%
\thanks{This work is developed within the framework of the Shuttle2X project, funded by the Federal Ministry for Economic Affairs and Climate Action (BMWK) and the European Union, under the funding code 19S22001B. The authors are solely responsible for the content of this publication.}
\thanks{$^1$Department of Technical Cognitive Systems, FZI Research Center for Information Technology, Haid-und-Neu-Str. 10-14, 76131 Karlsruhe, Germany, \texttt{\{orf, ochs, schotschneider, doll, heinrich, zofka, zoellner\}@fzi.de}}
\thanks{$^2$Karlsruhe Institute of Technology (KIT), Institute of Applied Informatics and Formal Description Methods, 76131 Karlsruhe, Germany, \texttt{marius.zoellner@kit.edu}}
}%

\IEEEpubid{\makebox[\columnwidth]{\hfill} \hspace{\columnsep}\makebox[\columnwidth]{ }}

\maketitle

\begin{abstract}
  Fault diagnosis is crucial for complex autonomous mobile systems, especially for modern-day
  autonomous driving (AD).
  Different actors, numerous use cases, and complex heterogeneous components motivate a fault
  diagnosis of the system and overall system integrity.
  AD systems are composed of many heterogeneous components, each with different
  functionality and possibly using a different algorithm (e.g., rule-based vs.
  AI components).
  In addition, these components are subject to the vehicle's driving state and are highly
  dependent.
  This paper, therefore, faces this problem by presenting the concept of a modular fault diagnosis
  framework for AD systems.
  The concept suggests modular state monitoring and diagnosis elements, together with a state- and
  dependency-aware aggregation method.
  Our proposed classification scheme allows for the categorization of the fault diagnosis modules.
  The concept is implemented on AD shuttle buses and evaluated to demonstrate its capabilities.
\end{abstract}

\section{INTRODUCTION}~\label{sec:introduction}
AD systems comprise heterogeneous software parts with different objectives, making development,
testing, admission, and operation, possibly with over-the-air updates, even more challenging.
In all these phases of an AD vehicle lifecycle, insights into the system are vital.
Developers must identify software deficiencies, authorities might require an interface to the
system's internals, and operators, regardless of whether they are private users or public
transport companies, must intervene quickly on the system's defects.
In addition, their expertise ranges from AD experts to average consumers without technical experience.
This demonstrates the need for a versatile interface to the AD system's internals.

Data is transformed and enriched in different consecutive parts of the system.
For example, the AD system pre-processes sensor data, extracts relevant features, and, based on
this, creates a solving strategy for a specific task.
Consequently, the various parts of the system are highly dependent on each other, leading also to
error propagation.
A fault diagnosis framework must consider these dependencies when providing insights into the
system's internals, e.g. by hiding irrelevant information from the user for clarity.
Moreover, such information can lead to fatal decisions. 

\begin{figure}[t]
  \centering
  \includegraphics[width=\columnwidth]{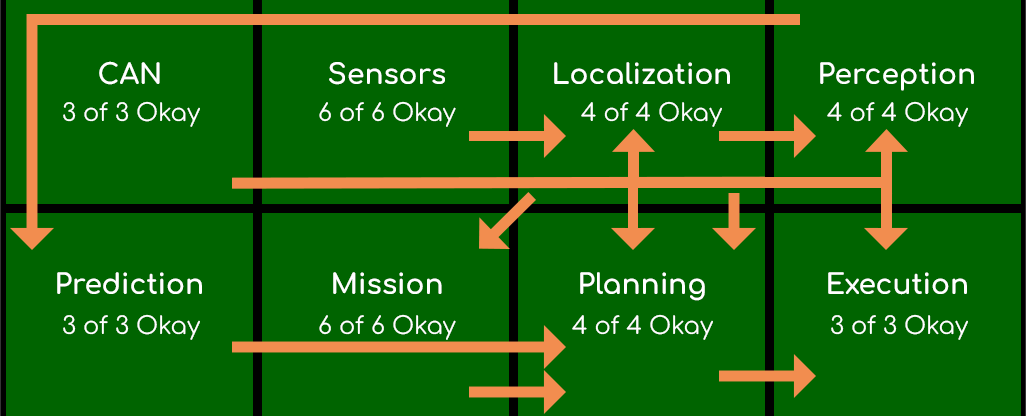}
    \caption{The modular diagnostic framework schematized here contributes to a system-wide diagnosis 
    by considering the diagnostic states of the submodules (e.g. "3 of 3 Okay") of each component 
    (green boxes, e.g. controller area network or "CAN") and their dependencies (orange arrows).
    The framework includes the current driving state of the AD vehicle, enabling the exclusion of
    irrelevant components from the diagnosis at a given time.}
   \label{fig:entry}
\end{figure}

Furthermore, the number of possible use cases will rise with the future application of AD 
vehicles.
AD is not only present in research anymore but is deployed in public spaces.
The driving task itself is increasingly moving into the background.
This shift from a sole driving task implies the system is in different states, like driving or
waiting at the bus stop.
Again, such conditions may influence the system's ability to build an adequate understanding of its
internals. 

Safety and robustness are still paramount objectives of an AD vehicle.  
A fault diagnosis that supports these objectives provides an understandable overview of its
functionality while considering the system's dependencies and states (Fig.\ref{fig:entry}).
Thus, we propose a modular fault diagnosis system for AD systems, which we describe in the
following.
After discussing the relevant literature in Sec.~\ref{sec:sota}, we briefly introduce the
terminology used throughout this paper in Sec.~\ref{sec:terminology}.
Then, we present our concept for a modular fault diagnosis system for AD in Sec.~\ref{sec:concept}.
We implemented this concept on our shuttles in Sec.~\ref{sec:implementation}, on which we also
conducted an evaluation (see Sec.~\ref{sec:evaluation}). Finally, we conclude our work in
Sec.~\ref{sec:conclusion}.

\section{STATE OF THE ART}~\label{sec:sota}

                   The failure-free functioning of a system is always a primary objective.
Foremost to ensure safety, but also to increase economic efficiency.
Especially as the complexity of systems in engineering fields increases, approaches to detect,
monitor, and prevent failures are vital.
As early as 1992, the importance of fault diagnosis was clear.
At this time, \cite{avizienis_dependability_1992} presented basic concepts, definitions, and
terminology of dependability as a "minimum consensus" within the "Reliable and Fault Tolerant
Computing" community.
Further works, e.g., from Isermann et al.
(\cite{isermann_trends_1997}) and van Schrick et al.
(\cite{van_schrick_remarks_1997}), shaped this field of science.
  Among others, newer works on general fault diagnosis approaches exist from Mazzoleni et al.
(\cite{mazzoleni_fault_2021}).
  Different surveys give an overview of the topic from an application perspective
  (\cite{treaster_survey_2004}, \cite{gao_survey_2015}).
There are also overviews of fault diagnosis in robotics (\cite{steinbauer_survey_2013},
\cite{khalastchi_fault_2019-1}), cyber-physical systems (\cite{yin_ieee_2020}), and automotive
systems (\cite{lanigan_diagnosis_nodate}).
In \cite{carlson_how_2005}, an analysis of the typical failures from crewless ground vehicles is
conducted.
It derives and examines statistics from ten different studies.
Sec.~\ref{sec:concept} of this paper provides a taxonomy and a fault categorization scheme.
While \cite{carlson_how_2005} also presents a fault classification, it differs from ours because it
is based mainly on the failure cause.

                   Many approaches for diagnosing faults in robotic or AD vehicle systems exist.
Thus, we only cite only a small selection.
The reader may refer to the surveys presented for in-depth research.
Already in 1997, neural networks were employed to implement a model-free fault diagnosis in
rigid-link robotic manipulators (\cite{vemuri_neural-network-based_1997}).
In \cite{boukhari_proprioceptive_2018}, recognition of failures in a vehicle's speed sensor is
presented.
Another approach tackles fault diagnosis of a steering actuator with a model-based support vector
machine classification (\cite{shi_fault_2021}).
We also refer to the author's work \cite{orf_modeling_2022} for detecting localization failures,
which integrates with the concept of this paper.
All these approaches have in common that they diagnose a particular part of the AD system rather
than providing a holistic application to the complete system.

                   An overall approach in robotics is presented in \cite{zaman_integrated_2013},
                   which leverages the robot operating system (ROS) \cite{quigley2009ros} and its
                   diagnostics stack to an integrated model-based diagnosis and repair
                   architecture.
This architecture builds up on essential elements, like observer modules.
We present similar modules here but with a more extensive framework, incorporating dependencies and
aggregation concepts.
Another more recent, overall framework for fault diagnosis in AD vehicles is presented in
\cite{min_fault_2023}, but focusing solely on sensor components.
In \cite{khalastchi_fault_2019}, an architecture for fault detection and diagnosis in robotics is
elaborated.
This architecture specializes in the typical components and system parts of robotics, e.g.,
sensors, perception, decision, and actuator components, while our concept is more versatile and
applicable to general complex systems.
We also refer to the author's previous work on a conceptual diagnostic system for AD
(\cite{orf_level_2020}), which forms the basis for this paper.

\section{Terminology}
\label{sec:terminology}
 We adopt the common terminology for faults and fault diagnosis, e.g.,
 \cite{avizienis_dependability_1992}, \cite{mazzoleni_fault_2021},
 \cite{isermann_supervision_2006}.
For detecting faults, the performance of a system has to be measured.
Its function must be defined to decide on a system or component's functionality.
This can be, for example, a simple threshold or more complex checks.
The system experiences a \emph{fault} if this defined property is not met.
A fault is not an event but a state or property of the system.
A faulty system has an \emph{error}, usually evident in incorrect output data.
But still, the system could fulfill its function with degraded performance, for example, if the
error is within a certain threshold.
A fault may further develop into a \emph{failure} of the system, meaning that it cannot fulfill its
required function.
A failure is an event resulting from single or multiple faults.
The different possibilities a system can fail are called \emph{failure modes}.
The \emph{failure cause} is the part of the system or the fault that led to the failure.
A system that detects faults and failures is called a \emph{diagnostic} or \emph{diagnosis system}.

\section{CONCEPT}~\label{sec:concept}

                   Fault diagnosis in AD systems has some peculiarities which distinguish it from
                   general fault detection.
A unique concept is required to address these specifics.

                   In this section, we first illustrate the preliminaries that lead to the specific
                   handling of fault diagnosis in AD systems
                   (Sec.~\ref{sec:concept_preliminaries}).
Our novel classification scheme of faults, which respects the preliminaries, is presented next
(Sec.~\ref{sec:concept_fault_classification}).
The concept utilizes diagnostic states, which we introduce in the following
(Sec.~\ref{sec:concept_error_states}).
We then describe concrete fault detection modules (Sec.~\ref{sec:concept_modules}).
Aggregation possibilities then complete the concept of fault detection in AD systems
(Sec.~\ref{sec:concept_aggregation}).

\subsection{Preliminaries} %
\label{sec:concept_preliminaries}
 The software for an AD vehicle is highly complex, as it is part of a distributed cyber-physical
 system with direct interaction with humans and their environment.
As such, the current actions of an AD vehicle impact the future.
Different components, e.g., perception, localization, prediction, planning, or execution,
raise various requirements, process multiple sensor data, and are built rule-based or data-driven.
This poses complex challenges to the operation of such a vehicle.
Errors and faults of the AD system need to be recognized and dealt with.
Also, the system needs to ensure reliability, safety, and security.
To not endanger lives, in or outside the vehicle, faults and errors must be eliminated, or, at
least, their impact has to be mitigated.
Faults can also propagate and lead to further, more severe faults and degrade the performance of
the vehicle.
Also, regulations and legal requirements expect such complex systems to be comprehensible,
especially in case of faults.
Having a fault recognition system gains the general public's trust in the safety of AD vehicles.
Furthermore, during development, a fault detection system can lead to a faster and better
improvement of this technology.
The necessity of a modular diagnostic system then arises for multiple reasons:
\begin{itemize}
  \item \emph{Scalability}: AD vehicles are subject of research.
        Their complex software system, comprised of distributed modules and components, evolves
        over time.
        AD systems' diagnostic part must also be modular to cope with such dynamic systems.
        On system changes or expansions, new modules should be introduced to the diagnostic system.
  \item \emph{Customization}: The requirements in a complex heterogeneous system are countless.
        A monolithic diagnostic system could not meet all these demands with ease.
        Modularity brings the possibility to tailor diagnostics to the specific needs of the
        vehicle system architecture.
  \item \emph{Fault Isolation}: Faults and errors can emerge from different parts of the system.
        One goal of the diagnostic framework is to identify the root cause of failures at a fine
        level of detail.
        Various diagnostic functions for different system parts may co-exist in a modular
        diagnostic system.
        Modular diagnostic units, thus, support the component- and function-specific isolation of
        faults, allowing for efficient troubleshooting and adequate use of countermeasures.
  \item \emph{Maintenance}: Especially throughout development, but also during operation, software
        updates can be necessary.
        Fixes to the software need to be manageable.
        A modular diagnostic system supports addressing updates to system parts independently and,
        thus, allows keeping the rest of the (diagnostic) system unchanged.
  \item \emph{Verification and Validation}: During development, verifying the
        correct implementation of the software components is essential.
        A modular diagnostic system aids this process.
        For vehicle admission, a validation of the AD system is helpful.
        Again, a modular diagnosis is indispensable, especially for validating the entire system,
        as it allows focusing on specific functions.
\end{itemize}
 An overall modular diagnostic system is essential to complex AD software.
In particular, modular diagnostics increases flexibility during operation and development and
ensures effective diagnostics, leading to safer and more reliable AD vehicles.

\subsection{Fault Classification}
\label{sec:concept_fault_classification}
\begin{figure*}[ht]
  \centering
  \includegraphics[width=\textwidth]{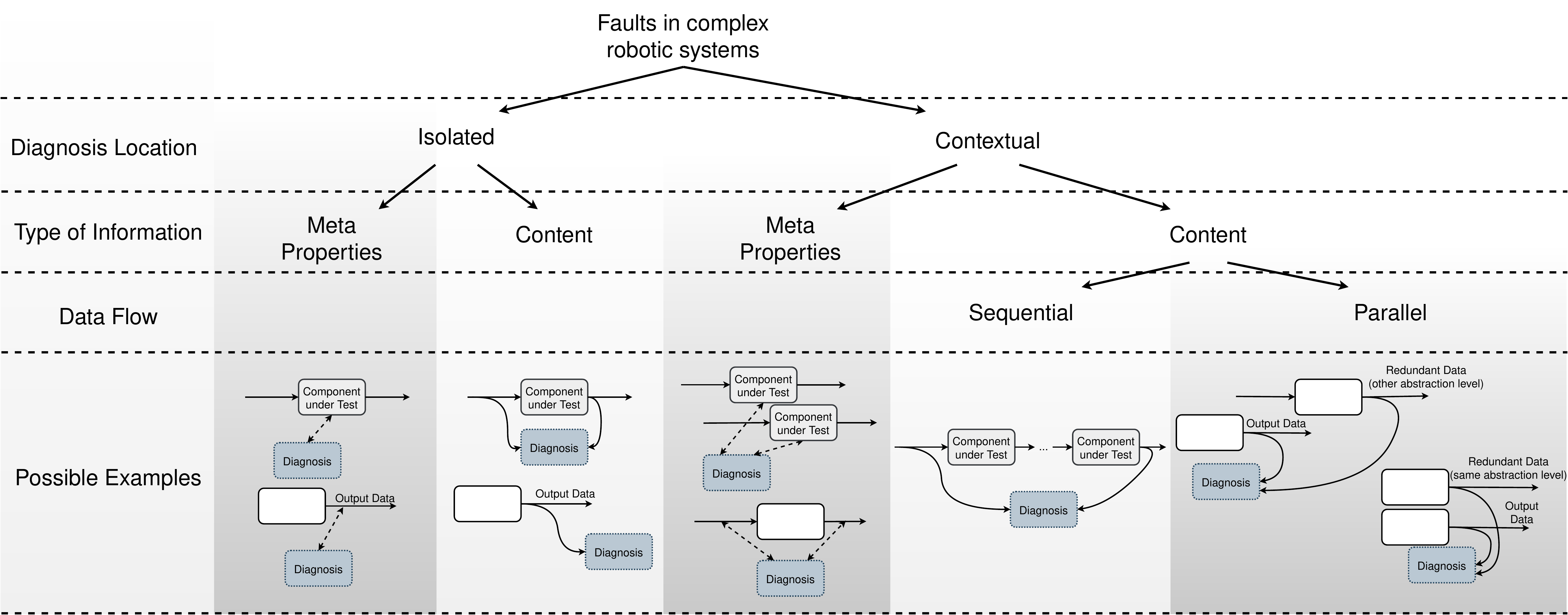}
  \caption{A general classification scheme for faults in AD, focusing on fault detection methods.
    The taxonomy is based on fault diagnosis location, the type of information processed, and the
    data flow.
    At the bottom, examples of the different categories are presented.
    The examples can roughly be grouped into general components and data stream diagnosis
    techniques.
  }
  \label{fig:classification_scheme}
\end{figure*}

 Fault detection in AD is challenging because of its complexity.
Software components with different objectives and technology, vast amounts of data processed, and
various hardware components on which the software components and data delivery are executed.
Besides a failure taxonomy in \cite{karapinar_robust_nodate}, which focuses on affected components
or methods, a general categorization for faults in AD systems is missing in the literature.
Also, a more general classification method based on fault detection methods is necessary.
Thus, we present a classification scheme for faults of software components in robotic systems
focusing on fault diagnosis location, type of information processed, and the flow of data
(Fig.~\ref{fig:classification_scheme}).

                   Complex AD software is often built using a modular component-based architecture
                   with message-passing data exchange.
The fault diagnosis should also reflect this by being modular, enabling a thorough fault analysis.
\begin{itemize}
  \item \emph{Diagnosis Location:}
        The question then arises of how many system parts are diagnosed by a single diagnostic
        module.
        For example, the diagnostic module could monitor a single component or data stream (see
        examples in columns one and two in Fig.~\ref{fig:classification_scheme}).
        This allows for an unclouded view of the system's elements but may be limited in diagnosing
        capabilities.
        By viewing multiple parts of the system, contextual information can provide a more profound
        diagnosis (examples in columns three to five in Fig.~\ref{fig:classification_scheme}).
        Of course, the validity of the diagnosis itself needs to be considered, as the context may
        also be faulty.
  \item \emph{Type of Information:}
        Diagnosing the system's parts capabilities should consider the information passed from
        component to component.
        In AD, the exchanged data often consists of sensor data, but arbitrary values are possible.
        All these types of information have in common that they can be used for fault diagnosis.
        A sensor might deliver incorrect data when faulty, for example.
        While this can provide precise insights into the functioning of the system's parts, there
        is also valuable meta-information available.
        Frequencies or latencies of message channels are just two examples that might indicate a
        fault.
        Thus, our taxonomy distinguishes the type of information processed for the diagnosis (see
        row two in Fig.~\ref{fig:classification_scheme}).
        Furthermore, this differentiation is present in isolated and contextual fault diagnoses.
        Generic examples can be found in columns one and three of
        Fig.~\ref{fig:classification_scheme}) and columns two, four, and five in
        Fig.~\ref{fig:classification_scheme}.
  \item \emph{Data Flow:}
        When diagnosing multiple system parts or data sources, the order of the information is
        relevant.
        Data is often enriched when passed from component to component.
        Diagnosis must consider this by comparing data from earlier stages with that from later
        system parts.
        We describe this as a sequential data flow (see column four in
        Fig.~\ref{fig:classification_scheme}).
        Although the processed data has a dependency relationship, such a diagnosis focuses on a
        distinct subsystem.
        Here, of course, fault-free earlier data is assumed.
        Whereas fault diagnosis based on independent data originating from different sources is
        denoted as parallel data flow (column five in Fig.~\ref{fig:classification_scheme}).
        This method can be more potent as other data types can be connected to reveal faults
        without assuming fault-free inputs, but it lacks focus on a distinct part of the system.
\end{itemize}
 Examples highlighting the taxonomy of fault diagnoses are depicted in
 Fig.~\ref{fig:classification_scheme} at the bottom.
Diagnosing meta properties of components or message channels are shown in the first column.
Meta-information could be message frequencies, latencies, or heartbeat signals from components.
Examples of the content level of isolated system parts in column two could be comparing components'
input and output data or evaluating message content.
The extension of fault detection through isolated to contextual meta-information (column three)
might be connecting such information from a component's input and output channels or from different
components.
When considering contextual content for fault diagnosis, an example of sequential data flow is
validating a chain of components' input and output data (column four).
One can imagine checking the output of an object recognition component (e.g., object lists) with
the input to a sensor data preprocessing component (e.g., raw sensor data) on which the object
recognition depends.
In contrast, an example of fault detection with parallel data flow (column five) could be redundant
information on different or the same abstraction layer, like object lists originating from separate
object detection methods or sensor data that do not depend on each other.

\subsection{Diagnostic States}
\label{sec:concept_error_states}
 The modularity of the fault recognition and diagnosis system makes a consistent representation of
 diagnostic states necessary.
A discrete model allows for easier comparison between heterogeneous components.
The degree of functionality of an element may be characterized differently depending on the applied
technique.
Thus, a mapping to a discrete diagnostic state is preferable.
Also, certain states that reflect more than a grade of functionality are beneficial.

                   We consider a classic ternary diagnostic state with additional non-functional states $\delta
                   = \{\textit{OK},\* \textit{WARNING},\* \textit{ERROR},\* \textit{IGNORE},$
                   $\textit{UNKNOWN}\}$ in our diagnostic concept.
The state $\textit{OK}$ indicates the normal and safe functioning of a system part, whereas
$\textit{ERROR}$ means a severe malfunction affecting the vehicle's safety and reliability.
To distinguish minor faults, we use the state $\textit{WARNING}$ to indicate abnormal behavior that
does not immediately lead to functional impairment and can be considered a pre-failure state.
We extend this by the states $\textit{IGNORE}$ and $\textit{UNKNOWN}$ to represent that a part of
the diagnosis is irrelevant or unknown.

\subsection{Modules}
\label{sec:concept_modules}
\label{sec:modules}
 The complexity of AD systems requires a modular diagnostic system
 (Sec.~\ref{sec:concept_preliminaries}).
The classification of faults (Sec.~\ref{sec:concept_fault_classification}) shows that different
types of fault diagnosis techniques exist.
We will now leverage this knowledge to develop specific fault diagnosis methods, regardless of
being done from within a system part or from the outside.
Diagnosis from within a component or self-diagnosis can be more accurate because internal
information is available.
Observing faults from the outside is more challenging, as only dedicated interfaces can be used.
On the other hand, self-diagnosis may be overconfident and suffer from the same faults as the
component itself.
The following methods are examples of self- and observer-diagnosis.
These are frequently encountered and, thus, universally applicable fault diagnosis modules:

\begin{itemize}
  \item \emph{Component's Self-State}: Components often provide a self-state because they have
        insights into their internal processing and processed data.
        The component can carry out a more detailed fault diagnosis with this information.
        An example of such a method is the accuracy that might be calculated during localization.
  \item \emph{Message Frequency/Latency}: Meta-information is a valuable source for fault
        diagnosis.
        The behavior of these system parameters tends to change when faults are present.
        Examples are frequency or latency measurements of message channels and response times of
        components.
  \item \emph{Single Value}: Message data does not always contain complex sensor data.
        Often, only single values are transmitted.
        Such data can be numeric or boolean values indicating a single system property.
        By applying thresholds to such values, the functionality of a system's aspect can be
        deduced.
  \item \emph{Component Watchdog}: Components can fail alltogether.
        While missing output data could identify a dead component, software or hardware components
        can usually also be probed over special interfaces.
        These components answer such requests with small responses on which the diagnosis can
        detect their basic reachability.
        Internet Control Message Protocol (ICMP) echo requests (commonly known as ping) are one
        example of a diagnosis based on meta-information.
  \item \emph{Domain-specific Diagnosis}: The most elaborated technique for fault diagnosis is a
        customized algorithm for a particular domain or component.
        Here, the diagnosis uses domain-specific knowledge to identify faults in components or
        message channels.
        Of course, these highly specialized methods must be tailored according to the use case.
        An example of this is a fault diagnosis for object detection that models the behavior of
        objects and validates the output messages against it.
\end{itemize}

\subsection{Aggregation}
\label{sec:concept_aggregation}
 The fault diagnosis modules presented in Sec.~\ref{sec:concept_modules} are each used for a
 distinct part of the complex system.
The diagnosis modules must be aggregated to detect faults in the complete system or larger
subgroups.
Furthermore, AD use cases are not only limited to the driving task.
An AD public transport vehicle undergoes different phases during driving, from initialization over
moving towards a goal to opening doors at stops.

Using the standard diagnostic states $\delta$ (Sec.~\ref{sec:concept_error_states}) among the modules
enables comparability and exchangeability.
Information from the fault diagnosis modules is grouped to reflect subparts of the system.
The aggregation of information can be done on multiple levels in a graph, with the
root ultimately representing the overall system.
The aggregation can be simple by just collecting fault diagnosis information from the individual
modules and deriving a collective state.
A group could be reported as $\textit{OK}$ if, and only if all submodules are $\textit{OK}$,
otherwise, the group would report $\textit{WARNING}$ or $\textit{ERROR}$.
In complex systems, this analysis of diagnostic states needs to be more elaborated.
Dependencies in the system are almost always present as information is processed by different
successive components, which all enrich or alter the data.
Faulty parts of the system are not only limited to a single component or message channel but
manifest in multiple areas of the system.

Arbitrarily detailed phases of the AD system are imaginable, posing challenges to fault diagnosis
aggregation.
Parts of the system may behave differently regarding fault diagnosis and should be handled
differently or excluded from the aggregation.

\section{IMPLEMENTATION}
\label{sec:implementation}
 In the preceding Sec.~\ref{sec:concept}, we outlined a general modular fault diagnosis
 architecture for AD systems.
This conceptual work is now implemented on our AD shuttles.
The shuttles are used mainly for research, i.e., developing and improving different aspects of
AD.
Nevertheless, the shuttles are also regularly employed to provide public transport.
The main advantage of our shuttles is the permit for driving in general traffic.
While driving in public, the law still requires safety-drivers in Germany.
In the public transport case, these are non-experts, making a comprehensible fault diagnosis
necessary.
In this section, we proceed with a specific implementation of the fault diagnosis in our AD shuttles.

\subsection{Aggregation}
\label{sec:aggregation}
\begin{figure}[th]
  \centering
  \includegraphics[width=0.65\columnwidth]{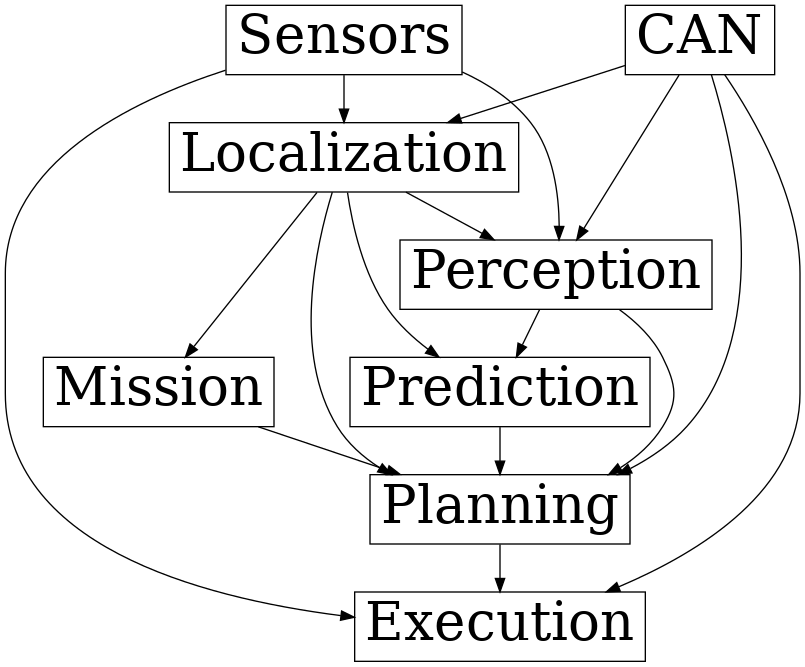}
  \caption{The dependencies of the aggregated fault diagnosis groups, corresponding to the high-level system parts.}
  \label{fig:agg_overview}
\end{figure}

\begin{figure}[th]
  \centering
  \vspace{0.2cm}
  \includegraphics[width=\columnwidth]{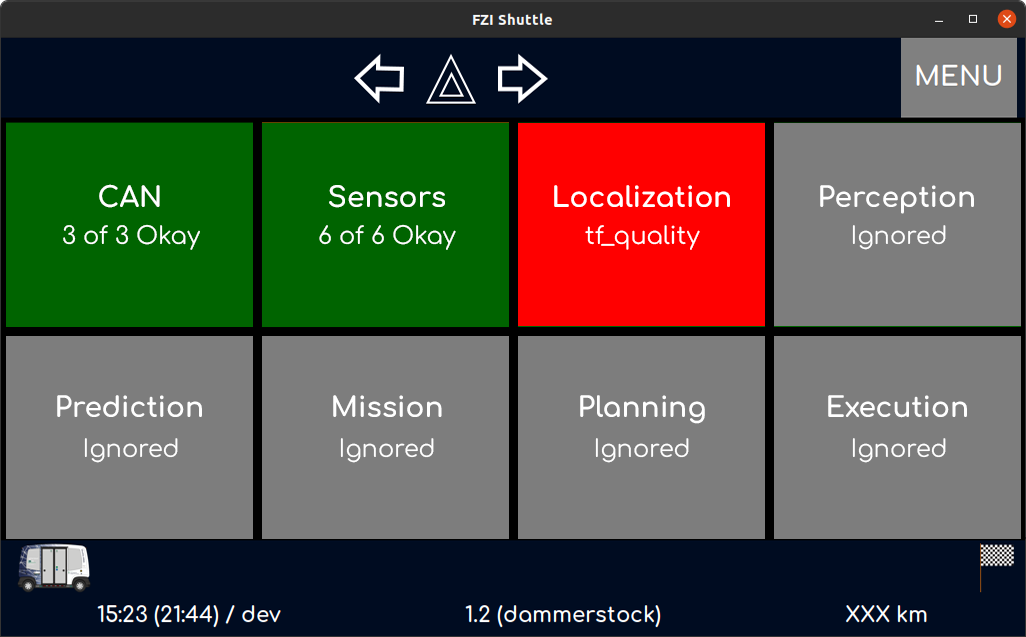}
  \caption{
    Detailed view of all aggregated fault diagnosis groups corresponding to the system's parts, displayed in the human-machine interface of the shuttles.
    Here, the \emph{Localization} group is in an $\textit{ERROR}$ state, resulting in dependent
    groups to be in the $\textit{IGNORE}$ state, depicted in grey with a label showing that they are ``Ignored''.
  }
  \label{fig:agg_detail_warning}
\end{figure}

 In this section, we present the aggregation of the individual modules from Sec.~\ref{sec:modules}
 and how these can represent a complex system.
Since every module works independently, they are unaware of the system or the state of other
components.
To incorporate the connection into a broader system, the aggregation is done through a generic
analyzer.
The analyzers employ a filtering mechanism to extract the essential diagnostic messages from the
influx of incoming data.
This filtering process is achieved by utilizing regular expressions and precisely defined
namespaces to identify and extract the required diagnostic information.
All information from the gathered modules defines the state of the aggregator.
It is even possible to have aggregators inside of aggregators.
Due to this fact, we can represent complex systems of multiple modules.
The missing link between components is the representation of dependencies, which the
dependency-aware analyzer does.
This specialized analyzer has multiple inputs which are observed.
If one of the submodules fails, this error is propagated further to the following dependencies.
The aggregated diagnostics and their corresponding dependencies are visualized in
Fig.~\ref{fig:agg_overview}.
A dependency is represented as edges in the graph.
A component can have multiple dependencies, exemplified by the \emph{Planning} and \emph{Execution}
components.
Conversely, a component can also serve as a dependency on multiple other components, as seen in the
case of the \emph{Sensors}.
In our paper, the focus is on the diagnostic graph employed for an AD function.
It is worth noting that this diagnostic framework can be readily modified to suit alternative
applications, such as service robots or control centers.
Due to knowledge of each aggregated component and its previous components, we can diagnose the root
cause by traversing the graph upon the faulty component.
This also improves the ability to display the responsible component and can be presented to an
operator (see Fig.~\ref{fig:agg_detail_warning}).

.

\subsection{Vehicle States}
\label{sec:vehicle_states}
 The implementation for our AD function encompasses diagnostics applications throughout various
 operational stages: start-up, idling, driving, and shut-down.
In each of these stages, not all modules are operational or initialized.
We utilize a state machine to monitor the vehicle's current status.

\begin{figure*}[th!]
  \centering
  \vspace{0.2cm}
  \includegraphics[width=\textwidth]{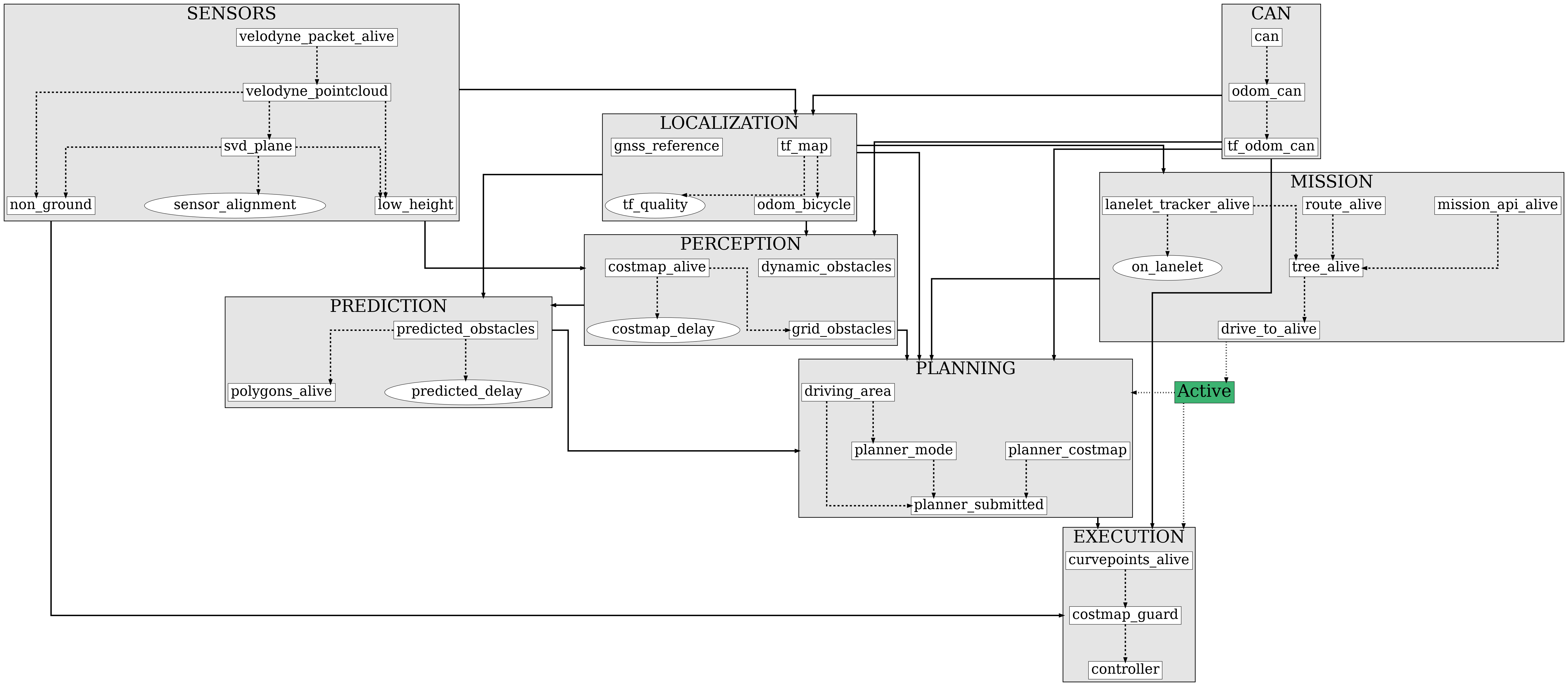}
  \caption{The dependency graph of the fault diagnosis modules with their aggregated groups (grey), representing high-level system parts, of our AD shuttles. Dependencies between fault diagnosis modules represent dependencies of the components of the AD
    system.
    The driving state of the vehicle (Active) is depicted in green. This state is set by the \textit{Mission} component and influences the behaviour of the \textit{Planning} and \textit{Execution} diagnosis modules.
  }
  \label{fig:agg_detail}
\end{figure*}

                   When the vehicle starts, the diagnostic system enters the default state with no
                   active diagnostic functions.
This precaution is taken to prevent unauthorized personnel from accessing any information.
Consequently, the initial step involves the safety operator logging in.
Upon successful login, the initial diagnostics phase commences, overseeing the incoming raw data
required for localization.
This includes the \emph{CAN} and \emph{Sensors}, specifically LiDAR sensors, as depicted in
Fig.~\ref{fig:agg_detail}.
Manual input from the safety operator is necessary to achieve successful localization, which is
required to initiate the processing of \emph{Perception} and \emph{Prediction} components.

                   The stage is reached to enable the AD function.
Except for \emph{Planning} and \emph{Execution}, all components are operational and running.
\emph{Planning} and \emph{Execution} components are activated upon receiving a mission and a goal position and
awaiting the clearance of the safety operator.
 With this, all diagnostics are running for a comfortable and safe ride.
During the whole ride, all components are diagnosed for proper operation.
If the diagnostics detects one component's $\textit{WARNING}$ or $\textit{ERROR}$ state, the
respective countermeasures are triggered (Sec.~\ref{subsec:countermeasures}).
Upon arrival, the diagnostics fall back to the default state for supervising all components except
\emph{Planning} and \emph{Execution}.
The switch for this is the \textit{Active} state (Fig.~\ref{fig:agg_detail}).

\subsection{Human Machine Interface}
\label{sec:hmi}
 In Sec.~\ref{sec:vehicle_states}, multiple inputs are needed from the safety operator to engage
 the AD function.
The safety operator is legally required to be in the vehicle at all times.
If the vehicle detects a false behavior, the built-in display enables a fast interface for the
safety operator.
The eight components \textit{Sensors}, \textit{Localization}, \textit{Perception},
\textit{Prediction}, \textit{Mission}, \textit{CAN}, \textit{Planning}, and \textit{Execution} are
distinguished (see Fig.~\ref{fig:entry} and Fig.~\ref{fig:agg_detail_warning}).
This division allows the safety driver to comprehensively overview the entire system and intervene
if necessary.

\subsection{Countermeasures}
\label{subsec:countermeasures}
 Recovering from possible component failures is crucial for safe AD and requires dealing with the
 failure, regardless of its severity.
We propose several functional and non-functional countermeasures to ensure the safety of the
current driving maneuver.
To avoid accidents due to the failure of essential maneuver components, we immediately decelerate
by \SI{1}{\meter\per\second\squared} if an $\textit{ERROR}$ state is present.
This action occurs when the maneuver component cannot perform a safe behavior, which would
incorporate static and dynamic objects.
A safer countermeasure is taken if the diagnostic module identifies component failures that are not
vital to the maneuver module.
The maneuver module will then compute a deceleration behavior considering static and dynamic
objects, leading to safer recovery.
Furthermore, the safety operator always sees the current diagnostic information of all components
on the Human-Machine-Interface (Sec.~\ref{sec:hmi}).
Component failures or warnings are shown immediately on the user interface.
In any case, we record data up to \num{30} seconds into the past to investigate the incident.

\section{EVALUATION} \label{sec:evaluation}

\begin{figure*}[t]
	\begin{center}
		\begin{subfigure}[T]{0.3\textwidth}
			\includegraphics[width=\textwidth]{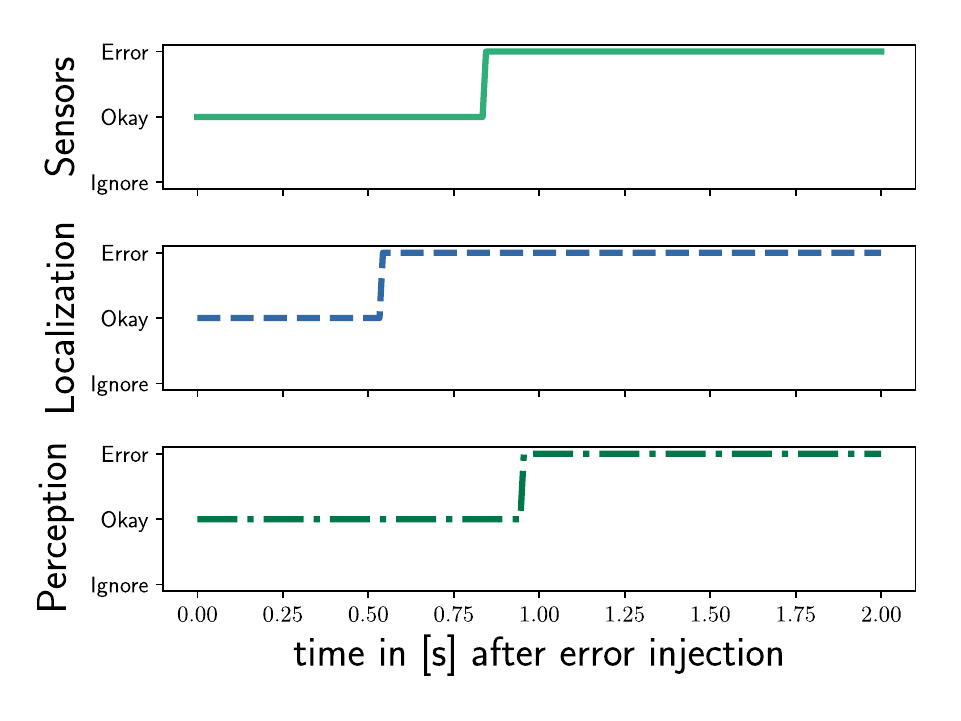}
			\caption{}
			\label{fig:scenario0}
		\end{subfigure}
		\begin{subfigure}[T]{0.3\textwidth}
			\includegraphics[width=\textwidth]{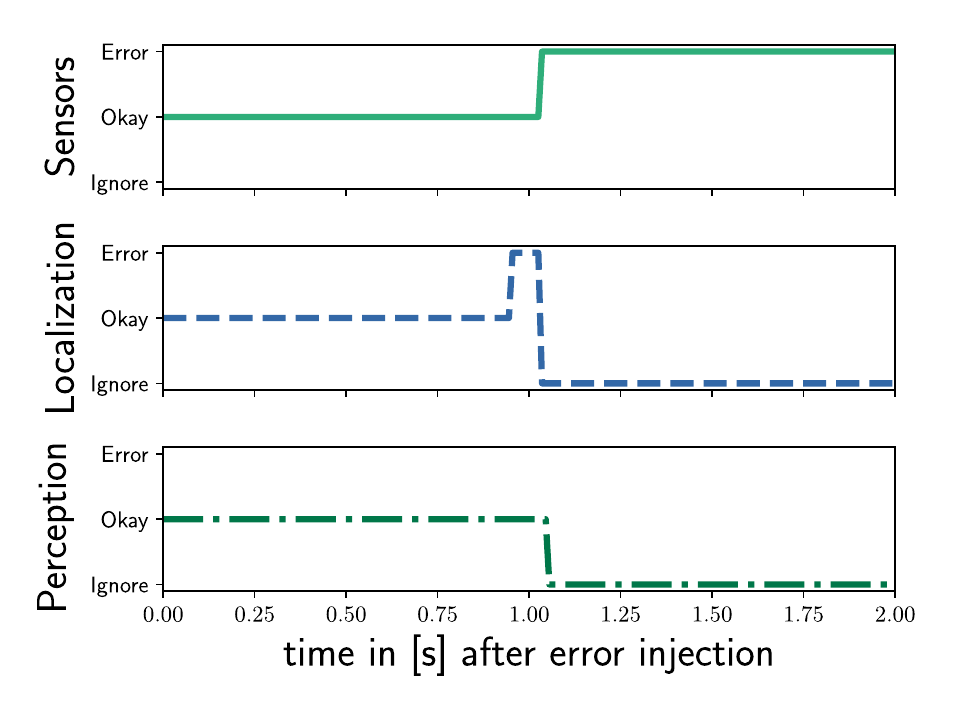}
			\caption{}
			\label{fig:scenario1}
		\end{subfigure}
		\begin{subfigure}[T]{0.3\textwidth}
			\includegraphics[width=\textwidth]{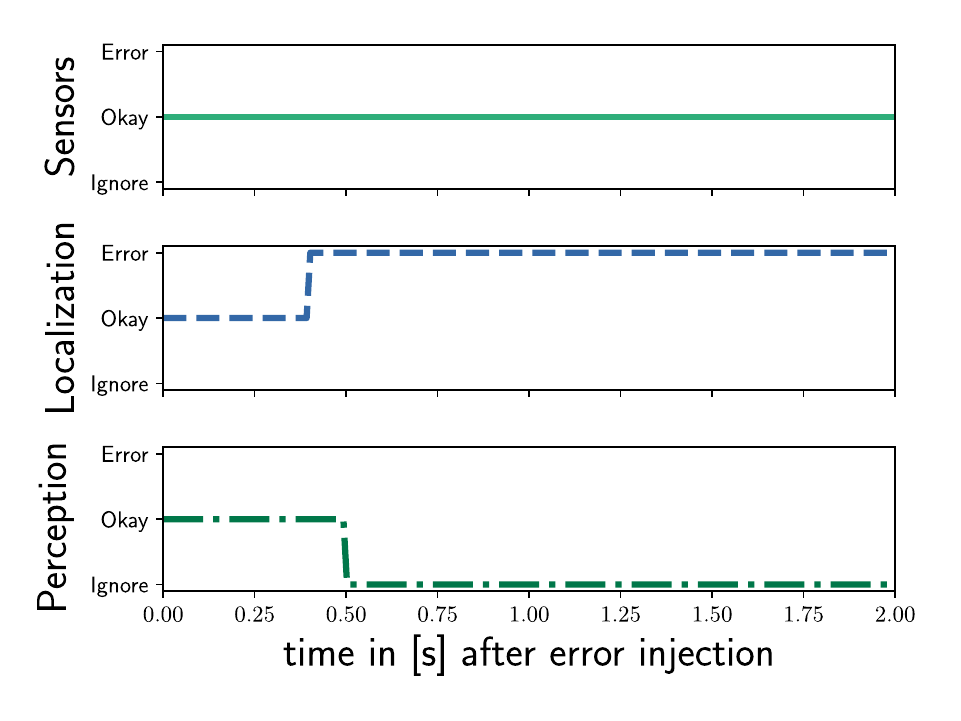}
			\caption{}
			\label{fig:scenario2}
		\end{subfigure}
	    \vspace{-0.1cm}
		\begin{subfigure}[T]{0.3\textwidth}
			\includegraphics[width=\textwidth]{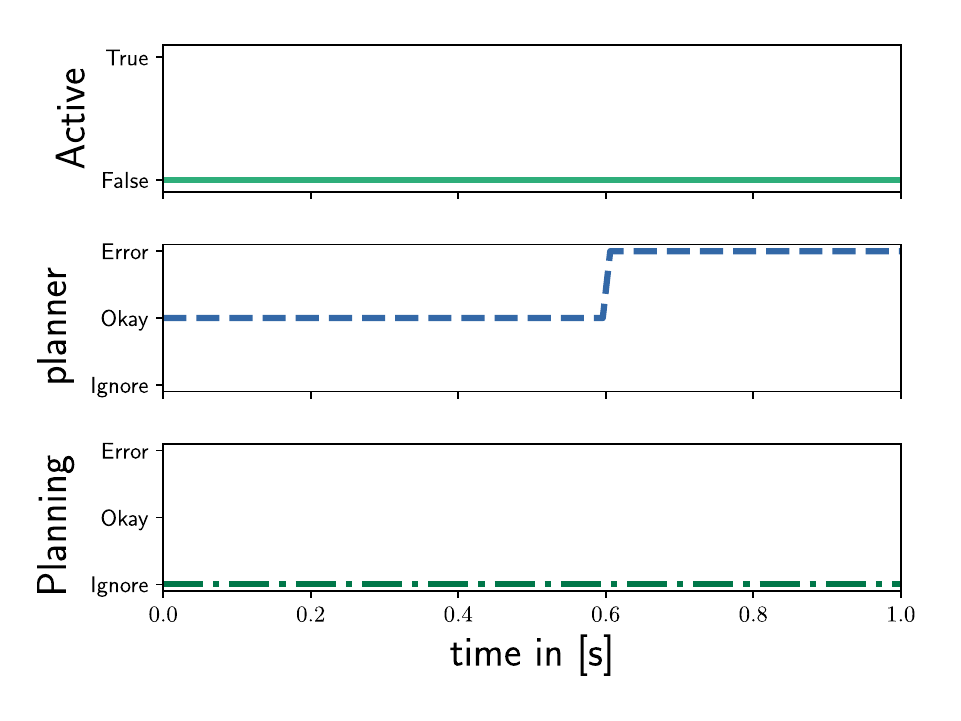}
			\caption{}
			\label{fig:scenario3}
		\end{subfigure}
		\begin{subfigure}[T]{0.3\textwidth}
			\includegraphics[width=\textwidth]{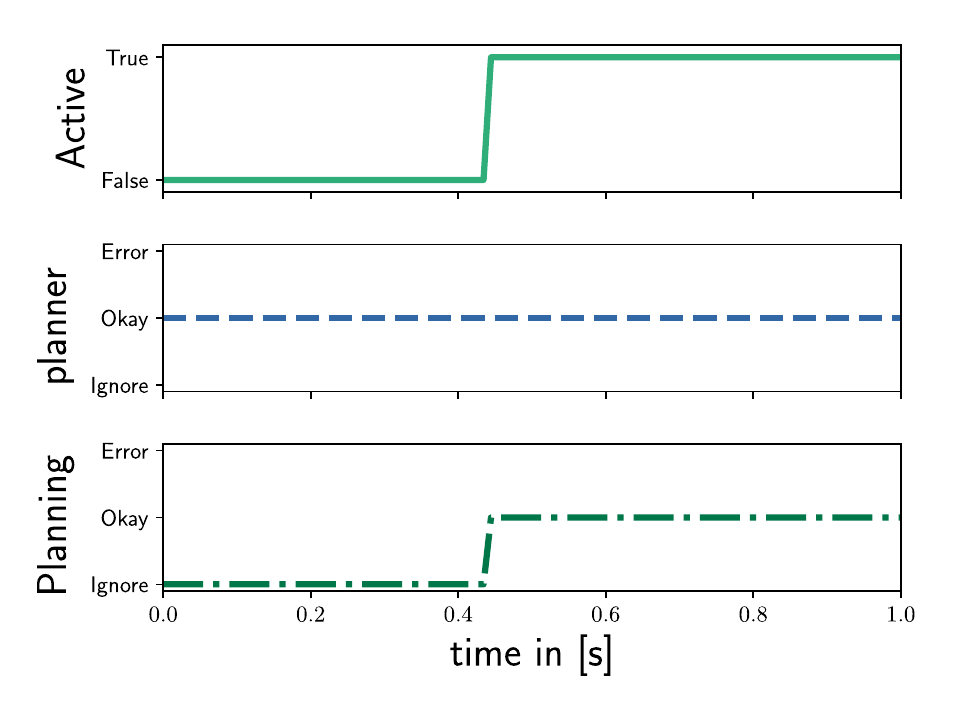}
			\caption{}
			\label{fig:scenario4}
		\end{subfigure}
		\begin{subfigure}[T]{0.3\textwidth}
			\includegraphics[width=\textwidth]{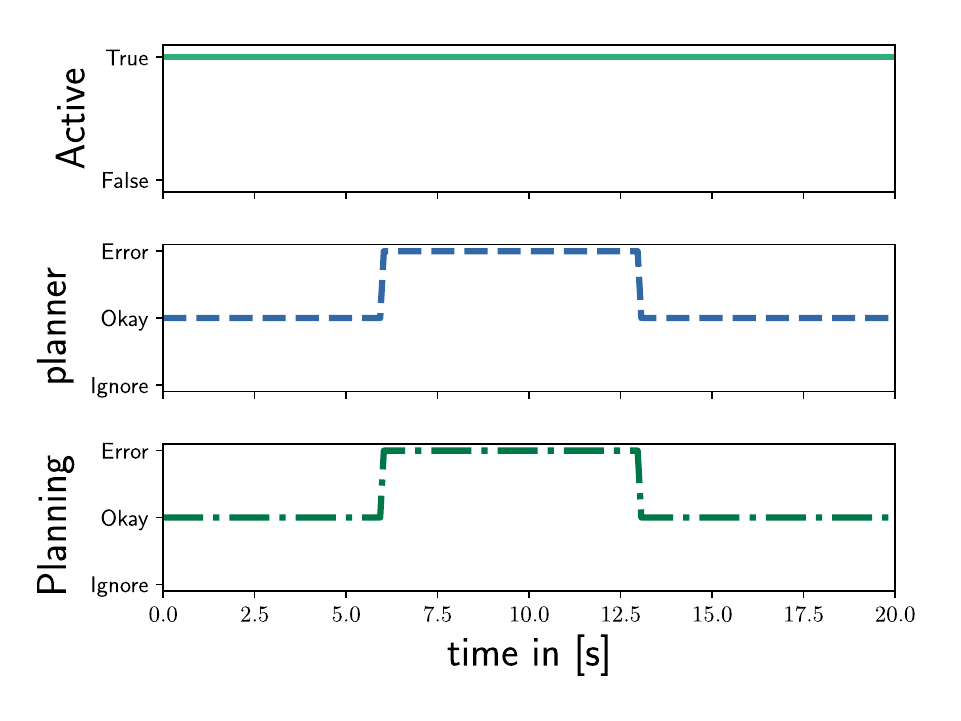}
			\caption{}
			\label{fig:scenario5}
		\end{subfigure}%
	\end{center}%
	\vspace{-0.1cm}
	\caption{Different aspects of fault detection during sensor outage (\ref{fig:scenario0} + \ref{fig:scenario1}), localization failure (\ref{fig:scenario2}), and planner problems (\ref{fig:scenario3}-\ref{fig:scenario5}).
		The top row shows the aggregated groups for the \emph{Perception} diagnoses and its dependent aggregated groups for the \emph{Localization} and
		\emph{Sensor} diagnoses and their reported fault states.
		The bottom row shows the fault states of the aggregated group \emph{Planning}, its failure
		module \emph{planner\_submitted} (here \emph{planner}), and the vehicle's driving state
		(\emph{Active}). Note the delay of the \emph{Sensor} fault state due to a sliding window approach, resulting in a later reporting.
	}
	\label{fig:full_dist_different_slam}
\end{figure*}

                   In this section, we demonstrate the capabilities of the modular fault diagnosis
                   implemented on our shuttles.
To show the different aspects of our fault diagnosis system, we split the evaluation into several
scenarios, depicted in Fig.~\ref{fig:full_dist_different_slam}.
In each scenario, failures of subsystems were induced.
Failures of the \emph{Perception}, \emph{Localization}, and \emph{Sensors} components are analyzed
in scenario 1 (Fig.~\ref{fig:scenario0}) to 3 (\ref{fig:scenario2}), whereas in scenario 4
(Fig.~\ref{fig:scenario3}) to 6 (Fig.~\ref{fig:scenario4}) failures of the \emph{Planning}
component is investigated.
In the shuttles, different fault diagnosis modules exist (see Fig.~\ref{fig:agg_detail}).
We limit the following illustration to certain modules for comprehensibility, although all modules
were involved in the evaluation, which is especially relevant for aggregation.
The examined modules are \emph{velodyne\_packet\_alive}, \emph{tf\_quality}, \emph{tf\_map}, and
\emph{costmap\_delay} for scenarios 1-3 and \emph{planner\_submitted}, together with the driving
state \emph{Active} for scenarios 4-6.
Each module is part of an aggregation group, representing a subsystem of the AD 
system (see Fig.~\ref{fig:agg_detail}).
The individual subsystems are \emph{Sensors} (\emph{velodyne\_packet\_alive}, \dots),
\emph{Localization} (\emph{tf\_map}, \emph{tf\_quality}, \dots), \emph{Perception}
(\emph{costmap\_delay}, \dots) and \emph{Planning} (\emph{planner\_submitted}, \dots).
Aggregation is done by applying a \emph{OR}-operation to the output of all diagnosis modules,
resulting in an $\textit{ERROR}$ state if at least one module reports $\textit{ERROR}$.
The same holds for \emph{WARNING}.
The \emph{velodyne\_packet\_alive} and \emph{planner\_submitted} modules are isolated
meta-information diagnoses on message channel frequency.
The same holds for the \emph{costmap\_delay} module, which measures message channel latency.
The \emph{tf\_map} module also diagnoses an isolated message channel but analyzes its content.
The \emph{tf\_quality} module, on the other hand, is a contextual diagnosis that analyzes the
content on the same abstraction level in a parallel fashion.
This module employs the localization diagnosis from \cite{orf_modeling_2022}.
The \emph{Active} state is \textit{true} if the vehicle is in AD mode and
\textit{false} if it is waiting for the next goal to be set.

                   In scenario 1 and 2 (Figs.~\ref{fig:scenario0}+\ref{fig:scenario1}) a sensor
                   failure is injected at $t=0$.
In scenario 1, without a dependency analysis, the fault is recognized by the different diagnosis
modules and their corresponding parent groups.
The failure exhibits a propagation effect throughout the system, impacting multiple diagnoses
modules, including those not directly linked to the sensor.
In scenario 2, only the \emph{Sensors} group shows an $\textit{ERROR}$ state.
The \emph{Localization} and \emph{Perception} groups are in a dedicated $\textit{IGNORE}$ state.
Note that the dependent \emph{Localization} group is shortly in an $\textit{ERROR}$ state as one
submodule recognizes the fault quicker.
The \emph{Sensors} group diagnosis modules employ a sliding window approach for frequency checks.
However, it takes some time to detect dropped messages and consequently decreases the frequency.
Scenario 2 demonstrates that our modular dependency-aware diagnostics approach allows for
identifying the directly affected component of a fault.
This is even more highlighted in scenario 3 (Fig.~\ref{fig:scenario2}).
Here, a failure in the \emph{Localization} system is injected.
Immediately, the corresponding $\textit{ERROR}$ state in the \emph{Localization} group is visible.
The dependent \emph{Perception} group is instantaneously put into $\textit{IGNORE}$ state, meaning a
parent in the dependency graph is in an $\textit{ERROR}$ state.
Note that the \emph{Sensors} group is unaffected by this behavior as it is not dependent on the
\emph{Localization}.

                   The state-aware functionality is shown in scenarios 4-6
                   (Figs.~\ref{fig:scenario3}-\ref{fig:scenario5}).
On the top in scenario 4, the \emph{Active} state is \textit{false}, meaning that the vehicle is
standing at a bus stop waiting for the next goal.
In this case, faults of the \emph{Planning} algorithm, represented by the \emph{planner\_submitted}
module (abbreviated as \emph{planner} in Figs.~\ref{fig:scenario3}-\ref{fig:scenario5}), do not
affect the \emph{Planning} group because a missing planning message is pointless in this state.
Thus, the $\textit{IGNORE}$ state is reported in the \emph{Planning} group.
Scenario~5 shows that, as soon as the system is \emph{Active}, the \emph{Planning}
group does not report the $\textit{IGNORE}$ state anymore.
And finally, in scenario 6, an $\textit{ERROR}$ in the \emph{planner\_submitted} module is
correctly aggregated into the \emph{Planning} group, which immediately shows an $\textit{ERROR}$.
This behavior demonstrates the capability of our modular diagnostics to be state-aware.
\section{CONCLUSION}~\label{sec:conclusion}
 This article presented a concept for a modular fault diagnosis framework.
This framework contains individual fault diagnosis modules, with which different aspects of the AD
system are monitored.
With the proposed classification scheme, the modules can be categorized with a focus on the fault
detection method.
A state- and dependency-aware aggregation of the individual modules allows for safety-supporting,
comprehensible fault detection of the entire system.
We implemented this concept on our AD shuttles and demonstrated its usefulness regarding
driving states and system dependencies.
With this framework, a fault diagnosis of a complex AD system can be conducted.
While the implementation of the proposed system on our AD shuttles involved a lot of expert
knowledge, future work might focus on automatically detecting the dependencies and states of the
diagnostic components.

\bibliographystyle{IEEEtran}
\bibliography{IEEEabrv,references}

\end{document}